\documentclass[10pt,twocolumn,letterpaper]{article}

\usepackage{iccv}
\usepackage{times}
\usepackage{epsfig}
\usepackage{graphicx}
\usepackage{amsmath}
\usepackage{amssymb}
\usepackage{multirow}
\usepackage{diagbox}
\usepackage{booktabs}
\usepackage[T1]{fontenc}

\newcommand\blfootnote[1]{%
\begingroup
\renewcommand\thefootnote{}\footnote{#1}%
\addtocounter{footnote}{-1}%
\endgroup
}

\usepackage[pagebackref=true,breaklinks=true,letterpaper=true,colorlinks,bookmarks=false]{hyperref}

\iccvfinalcopy 
\def\iccvPaperID{7261} 

\ificcvfinal\fi

\begin{document}

\title{Towards Inadequately Pre-trained Models in Transfer Learning}

\author{
\textbf{Andong Deng}\textsuperscript{1,$\dagger$}, 
\textbf{Xingjian Li}\textsuperscript{2,5,$\dagger$}, 
\textbf{Di Hu}\textsuperscript{3,*}, 
\textbf{Tianyang Wang}\textsuperscript{4}, 
\textbf{Haoyi Xiong}\textsuperscript{2}
\textbf{Cheng-Zhong Xu}\textsuperscript{5},
\\
\textsuperscript{1}University of Central Florida  \textsuperscript{2}Baidu Research \textsuperscript{3}Renmin University of China \\ \textsuperscript{4}University of Alabama at Birmingham \textsuperscript{5}University of Macau \\
\\
}


\maketitle
\ificcvfinal\thispagestyle{empty}\fi

\begin{abstract}
Transfer learning has been a popular learning paradigm in the deep learning era, especially in annotation-insufficient scenarios. Better ImageNet pre-trained models have been demonstrated, from the perspective of architecture, by previous research to have better transferability to downstream tasks\cite{Kornblith_2019_CVPR}. However, in this paper, we find that during the same pre-training process, models at middle epochs, which are \textbf{inadequately pre-trained}, can outperform fully trained models when used as feature extractors (FE), while the fine-tuning (FT) performance still grows with the source performance. This reveals that there is not a solid positive correlation between top-1 accuracy on ImageNet and the transferring result on target data. Based on the contradictory phenomenon between FE and FT that a better feature extractor fails to be fine-tuned better accordingly, we conduct comprehensive analyses on features before the softmax layer to provide insightful explanations. Our discoveries suggest that, during pre-training, models tend to first learn spectral components corresponding to large singular values and the residual components contribute more when fine-tuning.
\blfootnote{\noindent
\textsuperscript{$\dagger$}Equal contribution. Work is partly done when Andong was an intern at Baidu Research.
\textsuperscript{*}Corresponding author.
}
\end{abstract}

\section{Introduction}
Deep learning has achieved tremendous success in modern computer vision with the aid of the strong supervision of well-labeled datasets, such as ImageNet\cite{imagenet}. However, data annotation is notoriously labor-extensive and time-consuming, especially in some specific domains where 
expertise is highly required. In such scenarios, transfer learning is of great interest for practitioners to train deep models with a small labeled dataset. Fortunately, 
existing efforts observe that when training on large-scale datasets, middle features of DNNs exhibit remarkable transferability to various downstream tasks~\cite{zeiler2014visualizing,yosinski2014transferable}. This facilitates popular deep transfer learning paradigms of fine-tuning a pre-trained model (FT) or simply employing the pre-trained model as a feature extractor (FE). With relatively sufficient labeled examples, fine-tuning the whole network usually achieves higher performance. Despite this, FE is still important when training resources are limited, or end-to-end training is not feasible. For example, some applications combine DNN features and other handcrafted features to obtain both accurate and explainable shallow classifiers~\cite{lopes2017pre,rajaraman2018pre,rahman2021deep}. 

\begin{figure}[t]
    \centering
    \includegraphics[width=8cm]{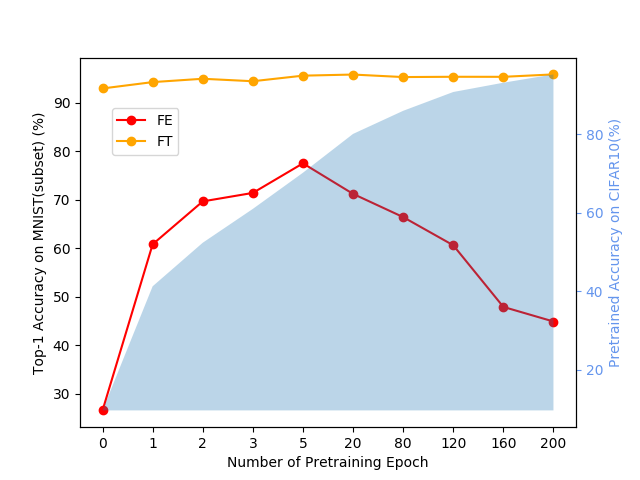}
    \caption{Toy experiment of transfer learning from a  ResNet18\cite{he2016deep} model pre-trained on CIFAR10\cite{krizhevsky2009learning} to a subset of MNIST\cite{lecun1998gradient}. FE means viewing the pre-trained model as a feature extractor, and FT means fine-tuning the whole model. It can be seen from the figure that the 5th-epoch model brings the best FE performance, which suggests that further pre-training on the source task would harm the feature quality for the target task. When fine-tuning the whole model, more adequate pre-training tends to deliver higher transfer learning performance.}
    \vspace{-0.1em}
    \label{toy experiment}
\end{figure}
 
Despite the ubiquitous utilization of pre-trained models, it still remains mysterious how such models benefit transfer learning.  
Several works pioneer to explore this plausible yet essential problem. 
\cite{Kornblith_2019_CVPR} systematically investigates whether better-performing models 
on source tasks, e.g. ImageNet, necessarily yield better performances on  
downstream tasks. They confirm this hypothesis for both FE and FT, over deep architectures with different capacities. However, recent works in the domain of adversarial training discover that an adversarially pre-trained model, though performs worse on ImageNet due to additional adversarial regularizations, can still transfer better than its natural (following the naming practice in \cite{utrera2021adversariallytrained}, referring to pre-training without adversarial methods) counterpart (with the same architecture)\cite{salman2020adversarially,utrera2021adversariallytrained}. In fact, these discoveries are 
to some extent in contradiction to the findings in \cite{Kornblith_2019_CVPR}, which argues that worse source models may transfer better. 

Our work investigates 
the influence of pre-training on transfer effects from a different perspective. Specifically, we focus on the trajectory of the pre-training process, inspired by recent studies on the learning order of DNNs. Several works~\cite{pmlr-v70-arpit17a,kalimeris2019sgd,li2019towards,mangalam2019deep} discover that DNNs tend to firstly learn simple and shallow features, e.g. colors and textures, 
which are regarded as more general and transferable across different data domains~\cite{zeiler2014visualizing}. From the perspective of the frequency domain, such features lie in low-frequency spectrums. On the other hand, several other works reveal that high-frequency features obtained by a pre-trained model are likely to cause a negative transfer~\cite{chen2019catastrophic}. 

The aforementioned observations motivate a question that,  does a fully pre-trained model definitely outperform its inadequately pre-trained version when transferring to target tasks (according to claims in \cite{Kornblith_2019_CVPR} ), or is there an intermediate pre-trained checkpoint that yields a better transfer effect than that of the fully pre-trained version? To our best knowledge, very little work manages to explore how the transferability of a model is impacted by the different stages in a pre-training process.

To investigate this question, we run a toy experiment using CIFAR10 as the source dataset and a subset of MNIST (we randomly choose 100 data points for each digit from the official training split, resulting in a 1000-sample training set) as the target. Briefly, we train a ResNet-18\cite{he2016deep} on CIFAR10 for 200 epochs and choose a set of checkpoints to run transfer learning in two different settings. In one setting, we treat the pre-trained model as a feature extractor (FE) and only retrain a softmax classifier, while in the other we fine-tune the whole model (FT). 
The retraining or fine-tuning continues for 100 epochs on the target dataset. 
As shown in Figure~\ref{toy experiment}, the best performance of FE comes from the early 5th-epoch model, while the FT performance is higher for later checkpoints. 

Two counter-intuitive facts can be observed from our results. One is that, a pre-trained model with higher accuracy on the source task is not necessarily better on the target task, especially when used as a feature extractor (FE). Among the checkpoints on the pre-training trajectory, there is no positive correlation between the source and target accuracy. The other observation shows inconsistent behaviors between FE and FT, indicating that a good starting point (FE) does not guarantee a good final result (FT). In order to explain the observed phenomenons, we investigate the spectral components of deep features before the FC layer (in Section~\ref{sec:SVD}), and observe that different parts of components contribute diversely for different pre-trained checkpoints within the same pre-training process.

In this paper, we conduct extensive transfer learning experiments,  including ImageNet and the other 9 benchmark datasets. The results suggest that, when retraining a new classifier on top of the features extracted from pre-trained models, inadequately pre-trained ImageNet models yield significantly better performance than that of the standard 90-epoch pre-trained version, but the performance still highly correlates with the source performance when fine-tuning. Further, we present insightful analyses to explain such a difference from the perspective of spectral components of the extracted features and find that there are specific components corresponding to pre-trained models at different pre-training stages. In summary, 
our main contributions are as follows:
\begin{itemize}
    \item Our work is the first to investigate how \textbf{different checkpoints} in the same pre-training process perform on transfer learning tasks. This contributes to a broader and deeper understanding of the transferability of neural networks. 
    \item We discover that in the same pre-training process, an \textbf{inadquately pre-trained model} tends to transfer better than its fully pre-trained counterpart, especially when the pre-trained model is used as a frozen feature extractor. We also further experimentally consolidate this claim beyond image classification.
    \item We observe that FT prefers later pre-training checkpoints, compared with FE. Our analyses based on spectrum decomposition indicate that the learning order of different feature components leads to different  preferences of pre-trained checkpoints between FE and FT. 
    \item We also point out the risk of utilizing transferability assessment approaches as a general tool to select pre-trained models. We evaluate LogME~\cite{you_logme:_2021}, LEEP\cite{nguyen2020leep} and NCE\cite{tran2019transferability}, which are dependent on frozen pre-trained models. Aiming to select the best pre-trained model among different checkpoints, scores obtained by these algorithms often show poor correlations with the actual fine-tuning performance.
\end{itemize}
\section{Related Work}
Pre-training on large datasets, such as ImageNet\cite{imagenet}, has long been a common method for transfer learning in various kinds of downstream tasks. 
Due to the huge effort brought by data annotation, researchers have reached a consensus that supervised or unsupervised pre-training as a parameter initialization or even an important medium for representation learning on existing large datasets is beneficial\cite{erhan2010does,jing2020self} for general downstream tasks\cite{noroozi2016unsupervised,misra2016shuffle,oord2018representation,tian2020contrastive,he2020momentum,chen2020simple,caron2020unsupervised,grill2020bootstrap,chen2021exploring,zbontar2021barlow} or specific ones\cite{xie2021detco,wei2021aligning,yang2021instance,li2023aligndet}. Zeiler et al. \cite{zeiler2014visualizing} have found that retraining a softmax classifier on top of a fixed pre-trained feature would benefit the classification of target data by a large margin compared with training from scratch. In recent years, designing different kinds of pretext tasks (e.g. jigsaw puzzle\cite{noroozi2016unsupervised}, rotation angle prediction\cite{gidaris2018unsupervised}, temporal order prediction\cite{misra2016shuffle}) as a self-supervised pre-training method became a popular trend in this community. Later on, contrastive learning\cite{oord2018representation,tian2020contrastive,he2020momentum,chen2020simple} has also been demonstrated as a better self-supervised pre-training approach. Beyond the vision domain, large-scale unsupervised pre-training in speech and natural language\cite{schneider2019wav2vec,devlin2018bert,yang2019xlnet,radford2019language,brown2020language} is appealing as well. Furthermore, learning universal representation and capturing cross-modal correspondence by pre-training in a multimodality setting\cite{radford2021learning,lu2019vilbert,li2022blip,li2023blip} and its downstream applications\cite{zhu2023minigpt,yu2023self} play an important role in the development of artificial general intelligence\cite{goertzel2007artificial}.

With such a powerful impact on deep learning, in the computer vision community, researchers have also been trying to understand the mechanism behind the success of pre-training, especially the ImageNet case, since ImageNet indeed has strong transferring power even to different data domains (e.g., in geoscience\cite{marmanis2015deep} and biomedical science\cite{raghu2019transfusion}). Erhan et al.\cite{erhan2009difficulty,erhan2010does} experimentally validated the role of unsupervised pre-training as a regularizer for the following supervised learning. Huh et al. \cite{huh2016makes}, via designing thorough experiments, answer a series of questions about the performance difference of transferring brought by different aspects (e.g., number of training samples, number of training classes, fine-grained or coarse-grained pre-training, etc.) of ImageNet. Using the proper normalization method and extending the training time, He et al.\cite{He_2019_ICCV} challenge this well-established paradigm and argue that it is possible to obtain better performance on target data from random initialization in detection and segmentation tasks. Following this work, Zoph et al. \cite{NEURIPS2020_27e9661e} further point out that self-training, with stronger data augmentation, can also lead to better transferring performance than pre-training. Nonetheless, pre-training is also viewed as a helpful training fashion for downstream tasks from different perspectives. Hendrycks et al.\cite{hendrycks2019using} have discovered that, in task-specific methods (e.g., label corruption, class imbalance, adversarial examples, etc.), pre-training enhances model robustness and brings consistent improvement compared with regular approaches. 

Aiming to investigate what kinds of pre-training models could bring better transferring performance, Kornblith et al.\cite{Kornblith_2019_CVPR} conduct extensive experiments on 16 different network architectures and suggest that models with higher top-1 accuracy on ImageNet could learn better transferable representations for target tasks. From the perspective of adversarial training, Utrera et al.\cite{utrera2021adversariallytrained} found that adversarially-trained models, though perform poorer on source data, actually have stronger transferability than natural models. And they further claimed that adversarially-trained models can learn more human-identifiable semantic information. Later, focusing more on model architecture, Salman et al.\cite{salman2020adversarially} drew the same conclusion, which further consolidates this viewpoint. In this work, we further investigate the relationship between top-1 accuracy on ImageNet and the transfer performance and found that some suboptimal models during pre-training transfer better when viewed as feature extractors, which is an analogous phenomenon with early-stopping\cite{prechelt1998early,yao2007early} in supervised learning that higher accuracy on training set does not mean higher test performance.

In order to further boost the performance of transfer learning, in several previous publications\cite{xuhong2018explicit,li2019delta,chen2019catastrophic}, new regularizers have been comprehensively investigated w.r.t. both model parameters and features. In \cite{xuhong2018explicit}, the convolutional weights are penalized to be closer to the source parameters rather than zero to avoid information loss from source data. Li et al. \cite{li2019delta} utilize an attention mechanism to restrict the difference between the convolutional features at the same hierarchy from the source model and target one, respectively. Further, Chen et al.\cite{chen2019catastrophic} claim that feature components corresponding to small singular values would be an impediment to knowledge transferring and then propose to suppress such components as regularization during fine-tuning. In this work, we also take advantage of Singular Value Decomposition on the features before the softmax layer and provide empirical analysis of the learning mechanism during the learning process.
\section{Experimental Setup}

We conduct extensive experiments on 8 representative natural image classification datasets (CIFAR10\cite{krizhevsky2009learning}, CIFAR100\cite{krizhevsky2009learning}, Food-101\cite{bossard14}, FGVC Aircraft\cite{maji2013fine}, Stanford Cars\cite{stanford_cars}, CUB-200-2011\cite{WahCUB_200_2011}, Oxford 102 Flowers\cite{nilsback2008automated} and MIT Indoor 67\cite{quattoni2009recognizing}) and one medical dataset (MURA~\cite{rajpurkar2017mura}) based on standard pre-training on both ResNet50~\cite{he2016deep} (90 epochs) and T2T-ViT\_t-14~\cite{yuan2021tokens} (300 epochs), which are representative architectures for ConvNets and Transformers in image classification. The top-1 accuracies are 76.06$\%$ and 81.55$\%$ for ResNet50 and T2T-ViT\_t-14, respectively. For pre-training details, we follow the standard ImageNet training configuration and the official T2T-ViT implementation for the two models, respectively.


\begin{figure*}[ht]
    \centering
    \includegraphics[width=0.32\linewidth]{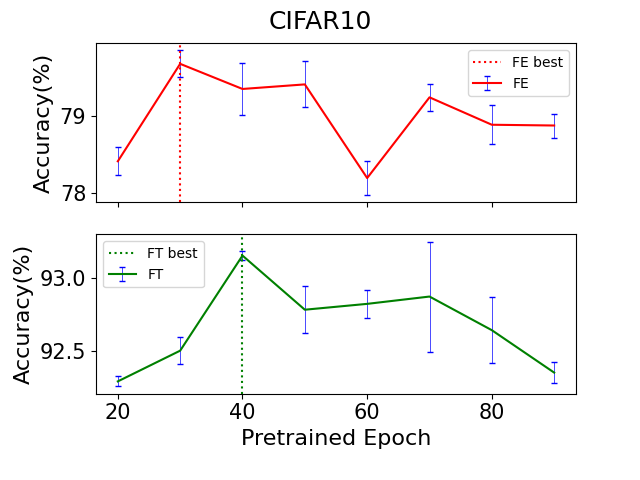}
    \includegraphics[width=0.32\linewidth]{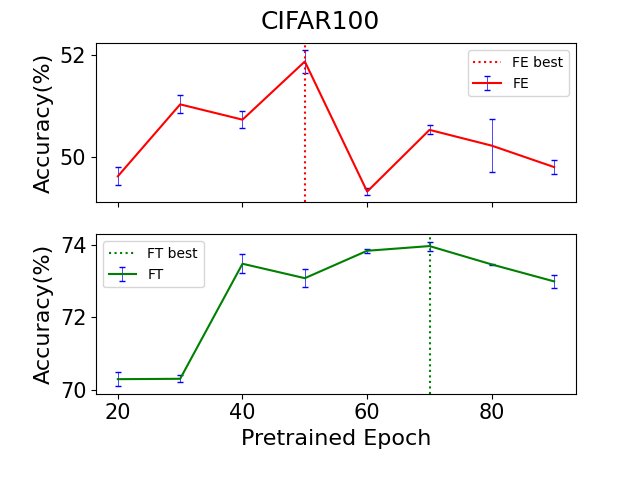}
    \includegraphics[width=0.32\linewidth]{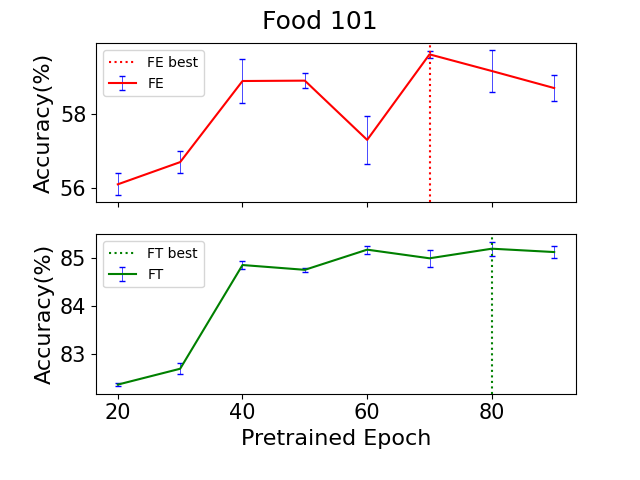}
    \includegraphics[width=0.32\linewidth]{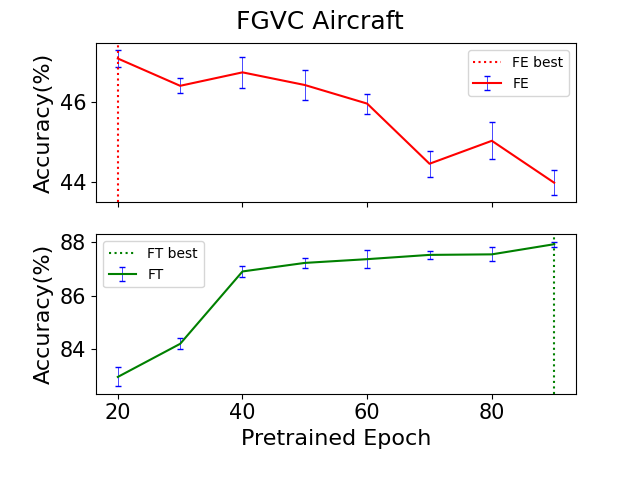}
    \includegraphics[width=0.32\linewidth]{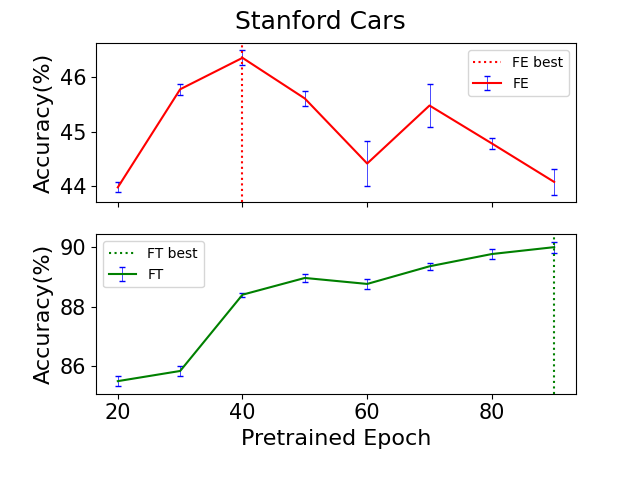}
    \includegraphics[width=0.32\linewidth]{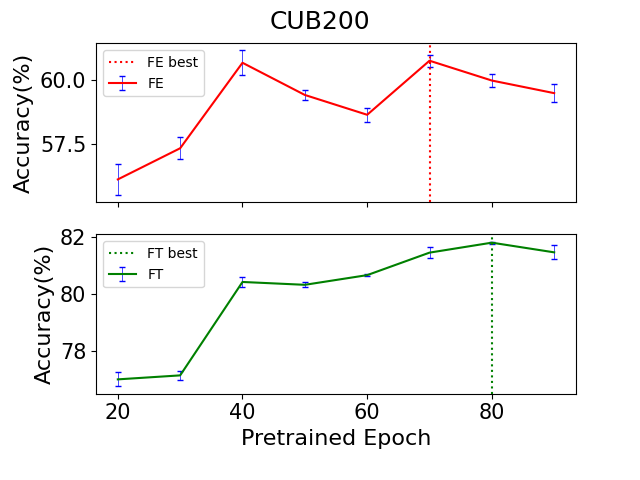}
    \includegraphics[width=0.32\linewidth]{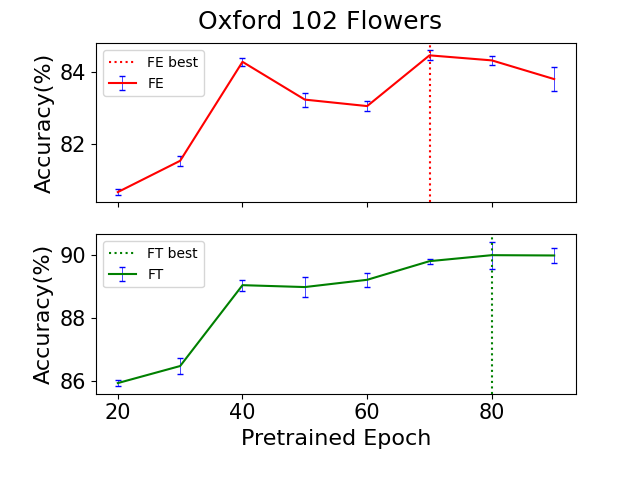}
    \includegraphics[width=0.32\linewidth]{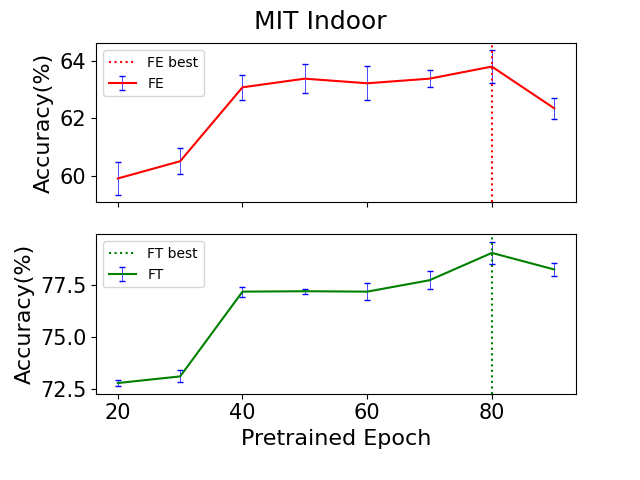}
    \includegraphics[width=0.32\linewidth]{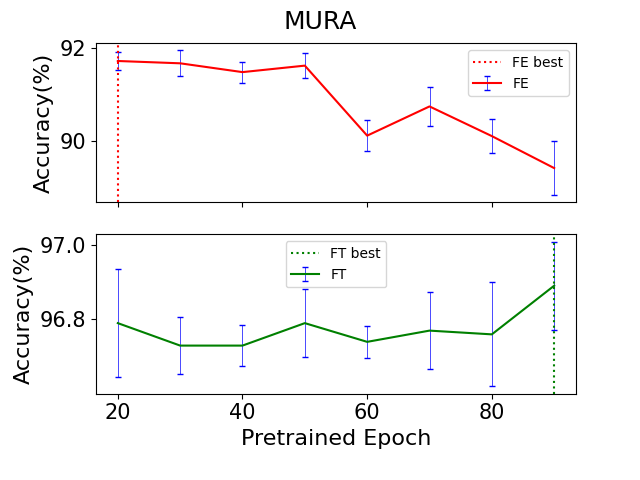}
    \caption{Transfer learning performance on selected  datasets. We can observe obvious different trends w.r.t. pre-trained epoch for FE and FT. The FT generally grows with the pre-training epochs increasing, while FE regularly reaches the peak at a middle epoch.}
    \label{fig:main_exp}
\end{figure*}

\begin{figure*}[t]
    \centering
    \includegraphics[width=0.32\linewidth]{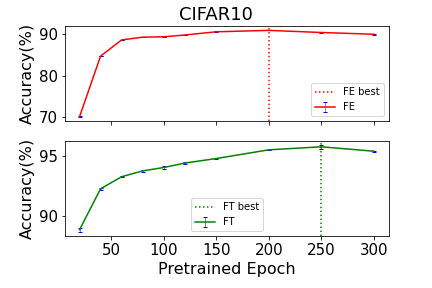}
    \includegraphics[width=0.32\linewidth]{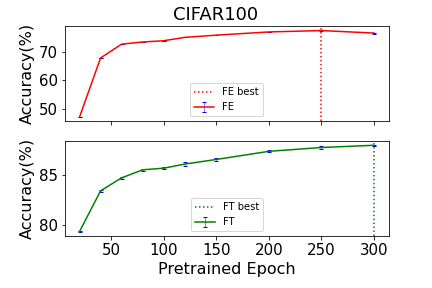}
    \includegraphics[width=0.32\linewidth]{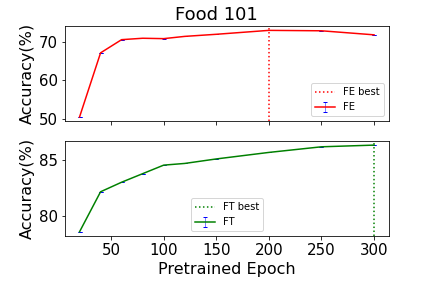}
    \includegraphics[width=0.32\linewidth]{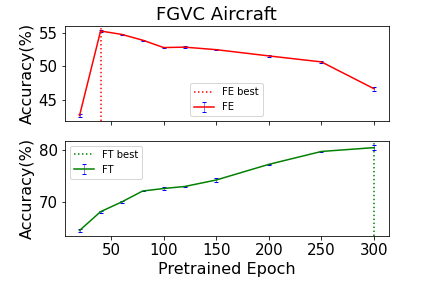}
    \includegraphics[width=0.32\linewidth]{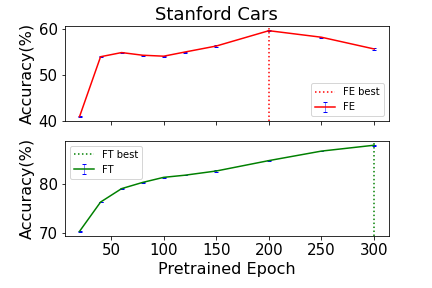}
    \includegraphics[width=0.32\linewidth]{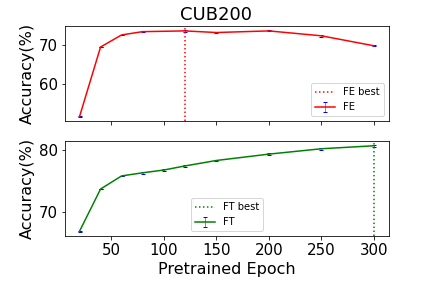}
    \includegraphics[width=0.32\linewidth]{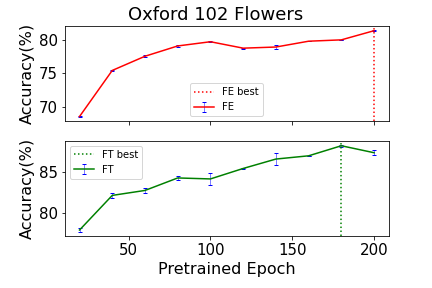}
    \includegraphics[width=0.32\linewidth]{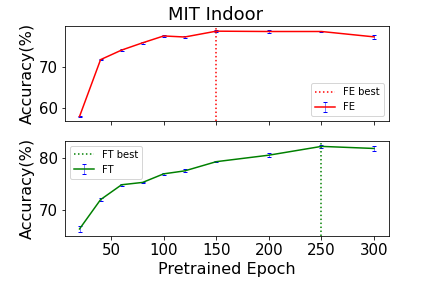}
    \includegraphics[width=0.32\linewidth]{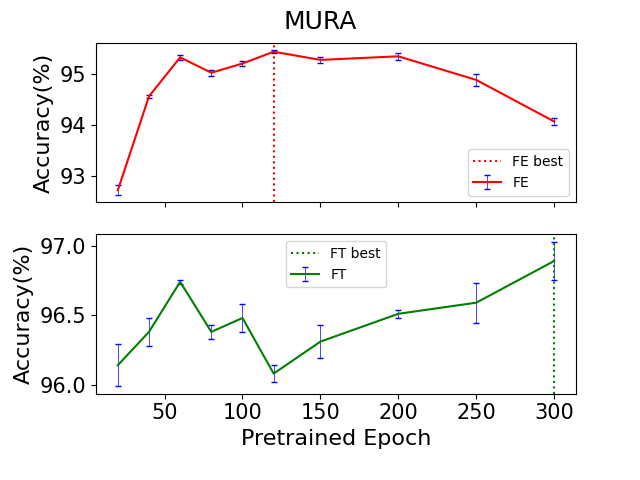}
    \caption{Transfer learning performance on selected  datasets on T2T-ViT~\cite{yuan2021tokens}. The trends are similar to those on ResNet50.}
    \label{fig:vit_exp}
\end{figure*}
\section{Results and Analyses}

In this section, we showcase all the experimental results of the transfer learning in two different settings: 1. Utilizing the pre-trained models as a feature extractor (FE) and retraining a softmax classifier; 2. Fine-tuning (FT) the whole model. We present experimental results of FE and FT in Section \ref{sec:fe} and \ref{sec:ft} respectively. A key observation is that inadequately pre-trained checkpoints transfer better for FE. 
Besides, we find that a better FE, which can be viewed as a better initialization for the target model, does not yield a better fine-tuning result. This is confirmed  in Section \ref{sec:tsne} by t-SNE\cite{van2008visualizing} visualization of deep features before the classifier. In Section~\ref{sec:SVD}, we manage to discover the in-depth learning mechanism during fine-tuning and empirically explain the aforementioned paradox, with the help of spectral components analysis.

\begin{figure}[ht]
    \centering
    \includegraphics[width=8cm]{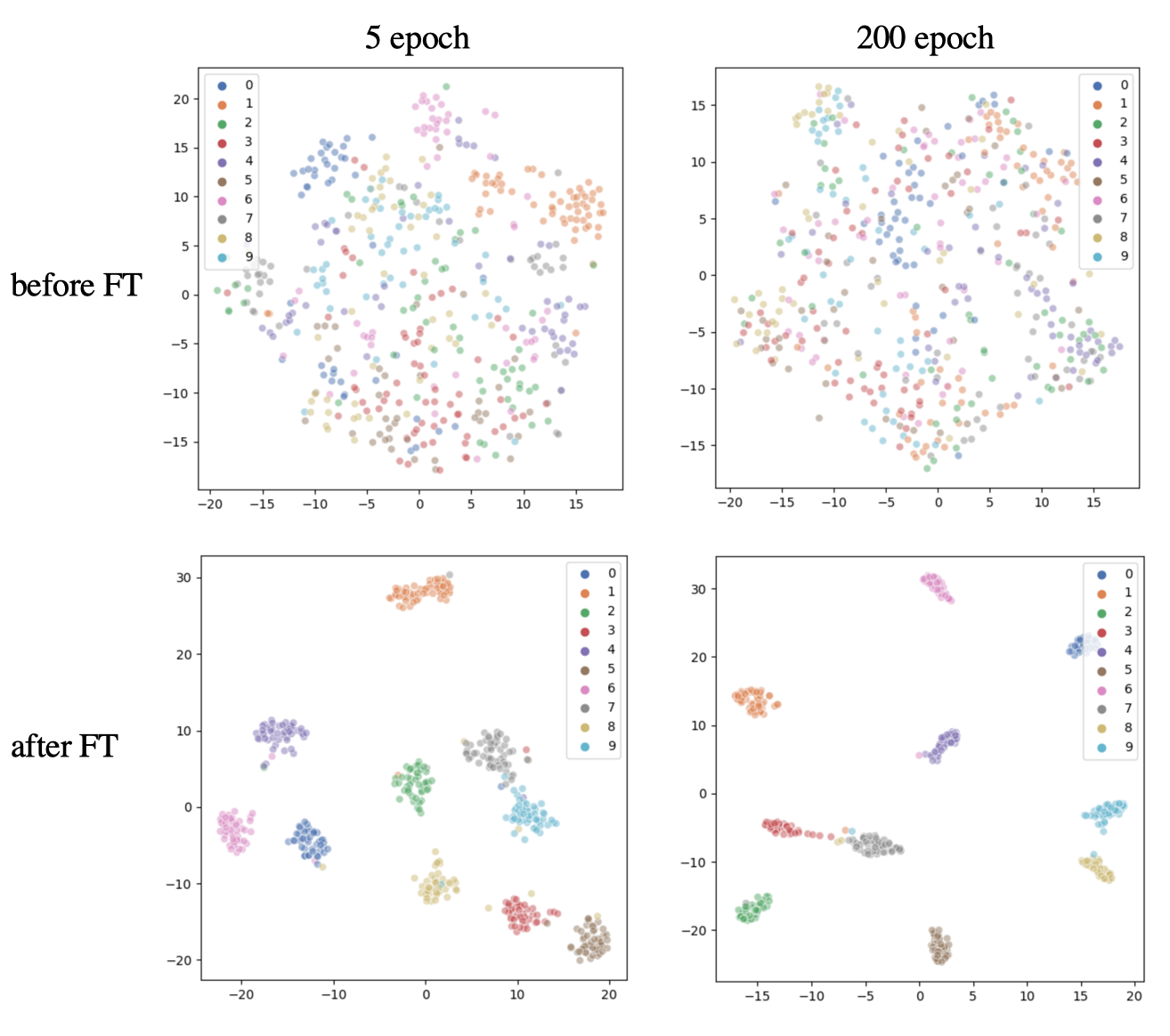}
    \caption{T-SNE visualization of features extracted before the classifier, before and after fine-tuning. Models are pre-trained on CIFAR10 and transferred to MNIST. The 5-epoch pre-trained model provides a better feature distribution on MNIST than the fully pre-trained one, but after fine-tuning for 50 epochs, the fully pre-trained model surpasses the 5-epoch one. Best viewed in color.}
    \vspace{-0.1em}
    \label{feature}
\end{figure}

\subsection{Inadequately Pre-training Brings Better Feature Extractors}
\label{sec:fe}
Concretely, in Figure \ref{fig:main_exp} and Figure~\ref{fig:vit_exp}, we can easily observe that there exists a best transferring spot before the model is fully pre-trained when viewed as a feature extractor for different datasets, which means that the correlation between the accuracy of the pre-trained model and the quality of the feature of the penultimate layer is not as positive as claimed in \cite{Kornblith_2019_CVPR}. We can also notice that the general trace of the FE performance is roughly a U-form curve with respect to the source performance, implying a potential trade-off between multiple factors during the pre-training process. 
Some curves exhibit a form of double-U, e.g. Stanford Cars and CUB-200-2011, and the FE performance at pre-trained epoch 40 and 70 is more likely to increase. We suspect this phenomenon may relate to the learning rate annealing after the 30-th and 60-th pre-training epoch~\cite{li2019towards}. In ~\ref{fe_app}, we will showcase some scenarios where inadequately pre-training brings advantages.

\subsection{Fully Pre-training Brings Better Fine-tuning Performance}
\label{sec:ft}
The case for FT is quite different compared with FE. The general evolution trend for fine-tuning is still positively correlated with the source performance, though the fully pre-trained checkpoint is not always the best. And we can also find that the best FT model emerges later than the best FE model. This asynchronization is actually surprising because it is common sense that a better initialization should bring better fine-tuning results. However, our work is not the only one that challenges this intuition. Recent empirical studies~\cite{zhang2020revisiting,li2020rifle} propose to improve fine-tuning by re-initializing top layers, i.e. employing a worse feature extractor as the starting point of fine-tuning.

\subsection{Visualization Analysis}
\label{sec:tsne}
In this subsection, we empirically visualize deep features of the best FE model (at the 5th epoch) and the fully pre-trained one (at the 200th epoch). The model is pre-trained on CIFAR10 and then transferred to MNIST, by both FE and FT. Deep features on the last convolutional layer of ResNet-18, produced by MNIST images, are extracted and dimensionally reduced to a 2-d space with t-SNE\cite{van2008visualizing}. The FE performances are 96.47$\%$ and 88.47$\%$, and the FT performances are 99.30$\%$ and 99.46$\%$(in this experiment we use the full version of MNIST). As can be seen from the top two plots in Figure~\ref{feature}, the visualization result is consistent with the transferring performance. 
When directly using the pre-trained model to extract features, data points in the embedding space of the 5-epoch model are clustered better, especially for categories corresponding to index 1 and 6; while the fully pre-trained model produces a more chaotic feature distribution that many data points are entangled with their incongruent neighbors. However, the situation becomes reversely when the whole model is fine-tuned. There exist a couple of misclassified data points in the feature space of the 5-epoch model, while the fully pre-trained model provides highly tight and discriminative features. This phenomenon is somehow surprising because this indicates that a better initialization, i.e., more discriminative features, might lead to worse fine-tuning performance.

\subsection{Spectral Component Analysis}
\label{sec:SVD}
Based on the observations from Figure~\ref{fig:main_exp} and Figure~\ref{fig:vit_exp}, two questions naturally arise: \textbf{What makes an inadequately pre-trained model a better feature extractor? What makes a better initialization (FE) perform worse than a fully pre-trained model which could not produce more discriminative features at the beginning?} To answer these questions, we resort to spectral analysis by Singular Value Decomposition (SVD) for an in-depth investigation. 
Specifically, we first obtain the batched feature matrix before the classification layer, which we denote as $\boldsymbol{F}\in{R^{b\times{d}}}$,
where ${b}$ is batch size and ${d}$ is feature dimension. After this, we decompose the matrix using SVD as:

\begin{equation}
\label{svd}
\begin{gathered}
\boldsymbol{F}=\boldsymbol{U}\boldsymbol{\Sigma}\boldsymbol{V^T},
\end{gathered}
\end{equation}
where ${\boldsymbol{U}}$ and ${\boldsymbol{V}}$ are left and right singular vectors respectively, and ${\boldsymbol{\Sigma}}$ is a rectangular diagonal matrix with the singular values on the diagonal. For convenience, we assume that all singular values are sorted in descending order.

Then we divide the diagonal matrix ${\boldsymbol{\Sigma}}$ as the main matrix ${\boldsymbol{\Sigma_m}}$ and the residual matrix ${\boldsymbol{\Sigma_r}}$. To achieve this division, we first calculate the sum over all singular values as $S^K_{\sigma} = \sum_{i=1}^{K} \sigma_i$, and then determine the minimum $k$ that satisfies $S^k_{\sigma} / S^K_{\sigma} \ge 0.8$. ${\boldsymbol{\Sigma_m}}$ preserves top $k$ lines of ${\boldsymbol{\Sigma}}$ and fills the remaining elements with zero. ${\boldsymbol{\Sigma_r}}$ is then obtained by ${\boldsymbol{\Sigma_r}} = {\boldsymbol{\Sigma}} - {\boldsymbol{\Sigma_m}}$. In this way, we can get two spectral components $\boldsymbol{F_m}$ and $\boldsymbol{F_r}$ of the original $\boldsymbol{F}$ by truncated SVD reconstruction as

\begin{equation}
\label{svd}
\begin{gathered}
\boldsymbol{F_m}=\boldsymbol{U}\boldsymbol{\Sigma_m}\boldsymbol{V^T}
\end{gathered}
\end{equation}
and
\begin{equation}
\label{svd}
\begin{gathered}
\boldsymbol{F_r}=\boldsymbol{U}\boldsymbol{\Sigma_r}\boldsymbol{V^T}.
\end{gathered}
\end{equation}

 According to \cite{chen2019catastrophic}, ${\boldsymbol{F_m}}$, as the main components of the feature matrix, represent the majority of transferring knowledge of the extracted features, while ${\boldsymbol{F_r}}$ is untransferable components or is hard to transfer that may do harm to the learning process and further causes negative transfer\cite{wang2019characterizing}.
 To evaluate the two components, we retrain a softmax classifier with Gaussian initialization on top of $\boldsymbol{F_m}$ and $\boldsymbol{F_r}$ for 50 epochs. We set the batch size as 128, using Adam\cite{kingma2014adam} optimizer, and the learning rate as 0.01. 
 
\begin{table}[t]
\fontsize{10}{10}\selectfont
    \caption{Results of Spectral Component Analysis for the best FE models and fully pre-trained one. SE denotes the pre-training epoch on CIFAR10, and SA means the pre-training accuracy. FE means viewing the pre-trained model as a feature extractor and only retraining a softmax classifier; FT means fine-tuning the whole model. The 96.47{$\%$} means the MNIST accuracy of the 5-epoch model in the FE task, and the 88.24{$\%$} means the classification accuracy on top of ${\boldsymbol{F_m}}$. We can observe that ${\boldsymbol{F_m}}$ and ${\boldsymbol{F_r}}$ perform differently no matter whether trained with more source information (pre-training) or more target one (fine-tuning): }
    \vspace{1em}
    \centering
    \label{svd_cifar_mnist}
    \scalebox{0.6}{
    \begin{tabular}{@{}c|ccc|ccc@{}}
    
    \toprule
    \diagbox[]{Task}{SE(SA)} & \multicolumn{3}{c|}{5 epochs (70.24{$\%$})}           & \multicolumn{3}{c}{200 epochs (95.32{$\%$})}                  \\ \midrule
    \multirow{2}{*}{FE}   & \multicolumn{1}{c|}{\multirow{2}{*}{\quad96.47{$\%$}\quad}} &  ${\quad\boldsymbol{F_m}\quad}$ & 88.24{$\%$\quad}  & \multicolumn{1}{c|}{\multirow{2}{*}{\quad88.47{$\%$}\quad}} &  ${\quad\boldsymbol{F_m}\quad}$ & 58.74{$\%$\quad} \\
                             & \multicolumn{1}{c|}{}                       &  ${\boldsymbol{F_r}}$  & 55.45{$\%$\quad} & \multicolumn{1}{c|}{}                       & ${\boldsymbol{F_r}}$ & 71.28{$\%$\quad} \\ \midrule
    \multirow{2}{*}{FT} & \multicolumn{1}{c|}{\multirow{2}{*}{99.30{$\%$}}} & ${\boldsymbol{F_m}}$ & 99.26{$\%$\quad} & \multicolumn{1}{c|}{\multirow{2}{*}{99.46{$\%$}}} & ${\boldsymbol{F_m}}$ & 99.08{$\%$\quad} \\
                             & \multicolumn{1}{c|}{}                       & ${\boldsymbol{F_r}}$ & 27.77{$\%$\quad} & \multicolumn{1}{c|}{}                       & ${\boldsymbol{F_r}}$ & 54.69{$\%$\quad} \\ \bottomrule
    \end{tabular}}
\end{table}
 
 For comparison, we choose the best FE model and the fully pre-trained model in this experiment. For convenience, we call the feature from the best FE model as BFE feature and the feature from the fully pre-trained model as FP feature. The first model pairs are from the CIFAR10-to-MNIST experiment, and the results are shown in Table~\ref{svd_cifar_mnist}. 
 Since we only analyze the features before the softmax layer, the FE models are actually identical to the corresponding pre-trained models; the FT models are fine-tuned with MNIST for 50 epochs. 
 The best FE model is the 5-epoch pre-trained model, whose accuracy is 96.47{$\%$} and is 8{$\%$} higher than the fully pre-trained one; however, after fine-tuning, the fully pre-trained model outperforms the 5-epoch one, even with less discriminative initial features.
 Thus, we decompose the BFE feature and FP feature to investigate which part of the components contributes to their higher performance in FE and FT, respectively. 
 
As can be seen from Table~\ref{svd_cifar_mnist}, there are several interesting discoveries as followed. 
\begin{itemize}
    \item \emph{The quality of $\boldsymbol{F_m}$ is responsible for the FE performance, while $\boldsymbol{F_r}$ is dominant when fine-tuning the whole model.} Specifically, we find that the 5-epoch model performs better as FE due to its remarkable superiority in $\boldsymbol{F_m}$. However, in the FT setting, the 5-epoch and 200-epoch models show similar performances in $\boldsymbol{F_m}$, and the higher $\boldsymbol{F_r}$ results in the higher overall performance of the 200-epoch model.
    
    \item \emph{As pre-training fits source data, $\boldsymbol{F_m}$ becomes less discriminative on  target data, but $\boldsymbol{F_r}$ transfers better} (observed from the line of FE in Table~\ref{svd_cifar_mnist}).  The degeneration in transferability of $\boldsymbol{F_m}$ could be caused by domain discrepancy between source and target data, as fully fitting source data may convert general patterns to those specific to the source domain. On the contrary, since $\boldsymbol{F_r}$ can not be well learned at earlier pre-training stages, it generally becomes  more informative by further pre-training. 

    \item \emph{For FT, $\boldsymbol{F_m}$ is easily adapted to target data, but $\boldsymbol{F_r}$ becomes less discriminative on target data} (observed from each column in Table~\ref{svd_cifar_mnist}). Both the 5-epoch and 200-epoch models achieve very high $\boldsymbol{F_m}$ performance (99.26$\%$ and 99.08$\%$) after fine-tuning. This implies the underlying learning mechanism that DNNs prefer a prior fitting with main spectrums rather than residual spectrums. The performance of $\boldsymbol{F_r}$ (from FE to FT) decreases due to the information capacity w.r.t. entire $\boldsymbol{F}$ is constant. Despite the degeneration, better $\boldsymbol{F_r}$ in FE still delivers better $\boldsymbol{F_r}$ after fine-tuning, indicating that the residual components learned from the source are not completely forgotten after fine-tuning on target.
    
\end{itemize}

\begin{figure}[t]
    \centering
    \includegraphics[width=8cm]{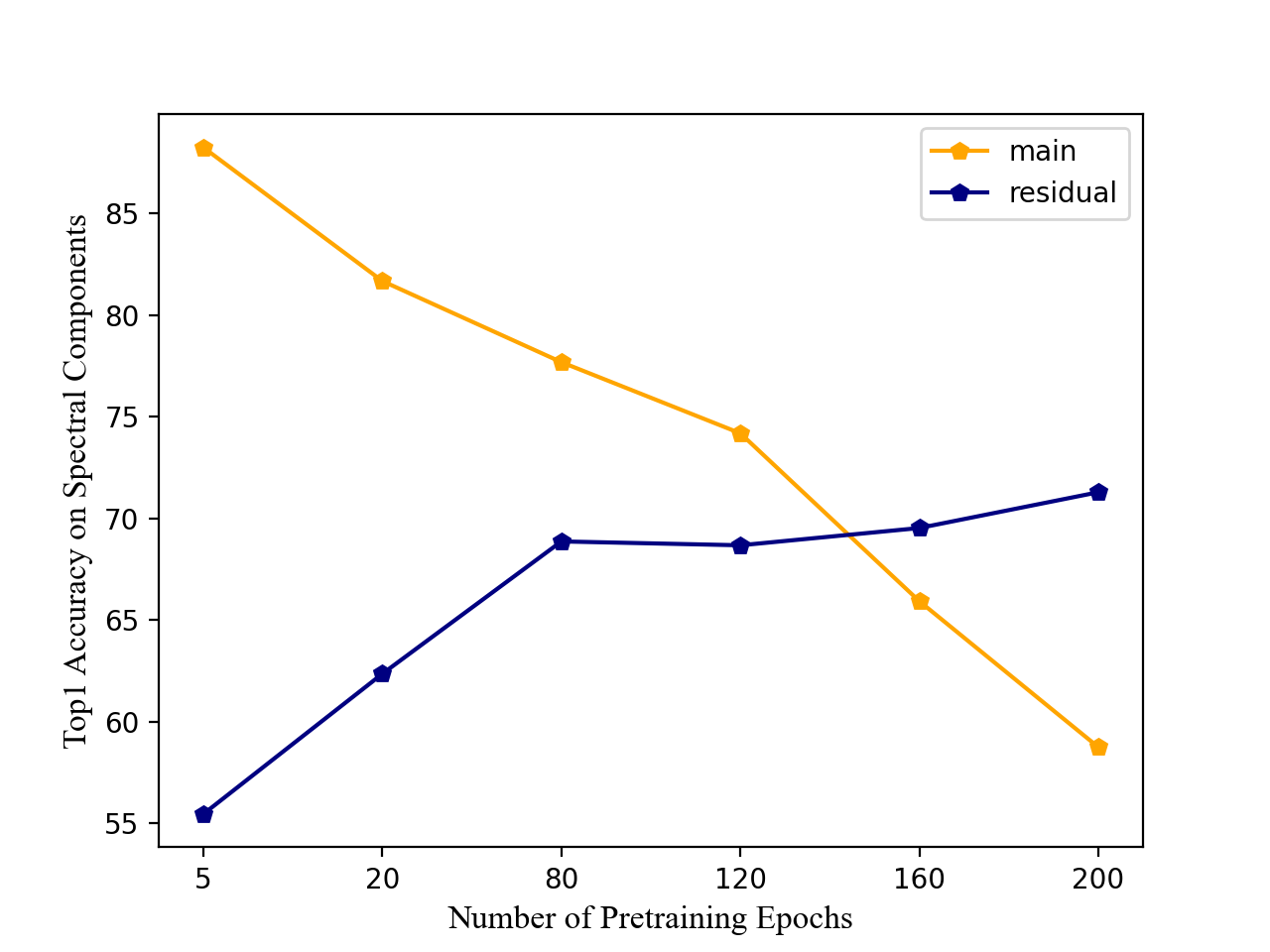}
    \caption{Evolution of the spectral components in pre-training from 5 epoch to 200 epoch. The orange curve represents the main components {$\boldsymbol{F_m}$}, and the blue one represents the residual {$\boldsymbol{F_r}$}. It is noticeable that $\boldsymbol{F_m}$ becomes less discriminative when the pre-training epoch grows. }
    \vspace{-0.1em}
    \label{fe evo}
\end{figure}

There might exist another explanation for the phenomenon in this spectral components analysis, which is from the perspective of the frequency domain. We can view $\boldsymbol{F_m}$ as low-frequency components of the original $\boldsymbol{F}$, and $\boldsymbol{F_r}$ as the high-frequency one. A couple of previous publications have revealed that the neural networks are inclined to learn low-frequency information first in the training process\cite{xu2019frequency,xu2019training,luo2019theory}. In our case, during pre-training, the model rapidly learns low-frequency knowledge at the early 5 epochs, which makes it the best feature extractor for downstream tasks. When keep learning in the source domain, more high-frequency patterns, which are specific to the source domain, are gradually learned; therefore, the negative transfer happens. 

We also illustrate the evolution of the classification performance of the two components for different pre-training epochs in the FE task (from CIFAR10 to MNIST) in Figure~\ref{fe evo}. It can be obviously noticed that $\boldsymbol{F_m}$ and $\boldsymbol{F_r}$ shows exactly opposite trends when pre-training epoch increases. With longer pre-training on CIFAR10, $\boldsymbol{F_m}$ becomes less discriminative since the model is prone to a deeper fitting to CIFAR10 with more high-frequency knowledge learned. Inversely, the residual components $\boldsymbol{F_r}$ becomes more informative for target data when memorizing more knowledge from the source domain.

\begin{figure*}[ht]
    \centering
    \includegraphics[width=0.32\linewidth]{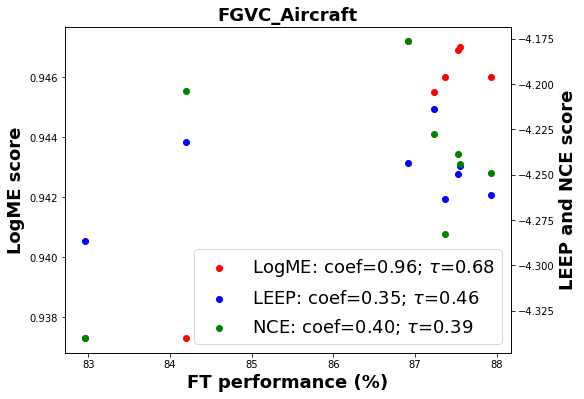}
    \includegraphics[width=0.32\linewidth]{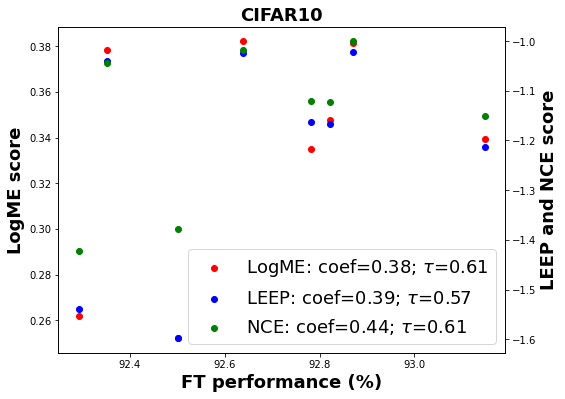}
    \includegraphics[width=0.32\linewidth]{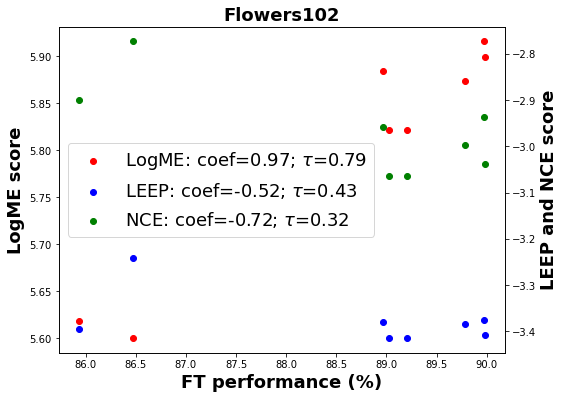}
    \caption{Transferability assessment for FGVC Aircraft, CIFAR10 and Flowers102 in and FT performance.}
    \label{fig:logme}
\end{figure*}

\subsection{Rethink Transferability Assessment Tools}

In this subsection, we utilize several transferability assessment tools LogME\cite{you_logme:_2021}, LEEP\cite{nguyen2020leep}, and NCE\cite{tran2019transferability} to validate whether it is possible to obtain the best checkpoint during pre-training without any training. We report these scores for three datasets at different pre-trained checkpoints and calculate the correlation coefficient and Kendall's $\tau$ coefficient\cite{kendall1938new} for FT performance. As can be seen in Figure~\ref{fig:logme}, LogME shows good ability in selecting the best checkpoint on FGVC Aircraft  and Flowers102, but a little bit poorer in CIFAR10; while the LEEP and NCE can hardly capture the correlation between the performance and the scores, especially in Flowers102.

\subsection{Application of Inadequate Pre-training}
\label{fe_app}
Pre-trained backbone networks are frequently used as feature extractors in downstream tasks, e.g., image captioning\cite{anderson2018bottom}, image retrieval\cite{song2019polysemous}, temporal action localization\cite{tian2018audio}, few-shot image classification\cite{qi2018low}, etc. In this section, we leverage typical downstream tasks to validate the effectiveness of inadequately pre-trained models, demonstrating the universal advantages of our method in various scenarios. 

We firstly focus on the image-text retrieval problem on the MSCOCO dataset\cite{lin2014microsoft} and choose the recent PVSE\cite{song2019polysemous} to incorporate our pre-trained ResNet50 models for evaluation. As shown in Table~\ref{tab:pvse}, we obtain the best retrieval performance with the 70-epoch pre-trained ResNet50, which extends our conclusion beyond image classification. We also provide a simple yet effective method in this case for selecting checkpoints. We use 25\% of the training data and the validation set to evaluate the models and select models according to rsum, which is the summation of the recall scores. The results in Table~\ref{tab:pvse_val} are consistent with that obtained when the full data is in use, demonstrating the rationale and efficacy of such a method. Moreover, to further consolidate our claim, we use our pre-trained ResNet50 models to perform a few-shot image classification task. Specifically, we choose weight imprinting\cite{qi2018low} on CUB-200-2011 in our experiment. In this method, the pre-trained ResNet50 models are firstly tuned on a subset of 100 classes, and then the classification weights are  imprinted to fit unobserved  classes. We directly take the accuracy of the 100-class subset as an indicator for model selection. In Table~\ref{tab:imprin}, we observe that  inadequately pre-trained models are still reliable feature extractors for this task. More importantly, it demonstrates that using the 100-class accuracy for model selection can also lead to the best 70-epoch model. 

\begin{table}[htbp]
\centering
\caption{Performance comparison of PVSE for different pre-training epochs. The 70-epoch pre-trained model obtain the best performance, which is consistent with our conclusion that inadequately pre-trained models could extract better visual features.}
\scalebox{0.88}{
\begin{tabular}{@{}lccccccr@{}}
\toprule
\multirow{2}{*}{ep} & \multicolumn{3}{c}{Image-to-Text} & \multicolumn{3}{c}{Text-to-Image} & \multirow{2}{*}{rsum} \\
\cmidrule(lr){2-4} \cmidrule(lr){5-7}
& R@1 & R@5 & R@10 & R@1 & R@5 & R@10 & \\ \midrule
50 & 9.88 & 28.28 & 41.26 & 3.84 & 12.58 & 19.88 & 114.98 \\
60 & 10.64 & 28.78 & 41.30 & 3.89 & 13.26 & 20.41 & 117.36 \\
70 & \textbf{12.28} & \textbf{33.46} & \textbf{45.82} & \textbf{5.20} & \textbf{16.46} & \textbf{25.44} & \textbf{138.66} \\
80 & 11.04 & 30.12 & 42.00 & 4.84 & 15.39 & 23.67 & 127.45 \\
90 & 10.82 & 30.70 & 44.50 & 4.58 & 15.18 & 23.63 & 132.67 \\
\bottomrule
\end{tabular}}
\label{tab:pvse}
\end{table}
\vspace{-3mm}

\begin{table}[htbp]
\centering
\caption{The best rsum score exists in the 70-epoch models.}
\scalebox{0.88}{
\begin{tabular}{@{}lccccccr@{}}
\toprule
\multirow{2}{*}{ep} & \multicolumn{3}{c}{Image-to-Text} & \multicolumn{3}{c}{Text-to-Image} & \multirow{2}{*}{rsum} \\
\cmidrule(lr){2-4} \cmidrule(lr){5-7}
& R@1 & R@5 & R@10 & R@1 & R@5 & R@10 & \\ \midrule
50 & 9.28 & 26.64 & 28.82 & 3.84 & 12.44 & 19.80 & 110.82 \\
60 & 9.64 & 28.26 & 39.88 & 3.82 & 12.91 & 20.410 & 114.91 \\
70 & \textbf{12.22} & \textbf{31.38} & 43.72 & \textbf{4.86} & 15.60 & 24.22 & \textbf{132.00} \\
80 & 11.60 & 31.38 & \textbf{43.98} & 3.54 & 12.06 & 19.52 & 121.87 \\
90 & 10.64 & 29.94 & 42.32 & 4.82 & \textbf{16.36} & \textbf{25.06} & 129.13 \\
\bottomrule
\end{tabular}}
\label{tab:pvse_val}
\end{table}
\vspace{-3mm}

\begin{table}[h]
\centering
\caption{Results of a classification task for different pre-training epochs. The table shows the performance for N-shot unseen classification and the accuracy on 100 seen classes.}
\label{tab:imprin}
\scalebox{1}{
\begin{tabular}{@{}lllllll@{}}
\toprule
ep & N=1 & N=2 & N=5 & N=10 & N=20 & 100-class \\
\midrule
50 & 52.07 & 55.64 & 60.77 & 63.95 & 65.64 & 81.28 \\
60 & 51.28 & 55.68 & 61.11 & 63.86 & 65.95 & 82.16 \\
70 & \textbf{52.92} & \textbf{57.44} & \textbf{62.20} & \textbf{65.08} & 67.05 & \textbf{82.82} \\
80 & 52.36 & 56.44 & 61.77 & 64.58 & 66.66 & 82.68 \\
90 & 52.11 & 57.34 & 61.49 & 64.83 & \textbf{67.17} & 82.47 \\
\bottomrule
\end{tabular}}
\end{table}
\vspace{-3mm}

\section{Discussion and Future Work}

\begin{figure*}[ht]
    \centering
    \includegraphics[width=0.32\linewidth]{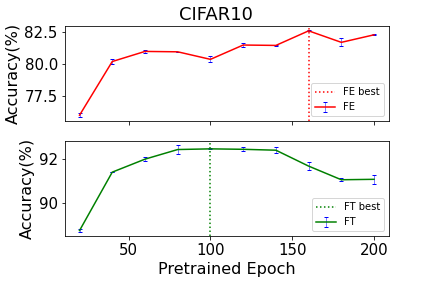}
    \includegraphics[width=0.32\linewidth]{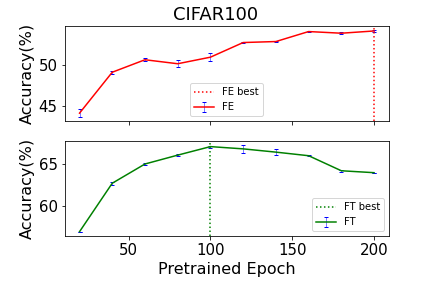}
    \includegraphics[width=0.32\linewidth]{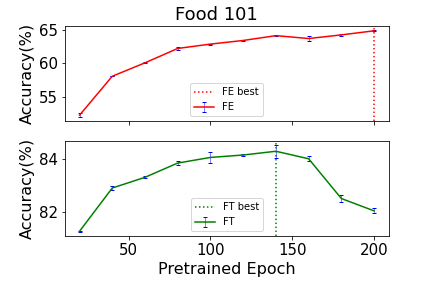}
    \includegraphics[width=0.32\linewidth]{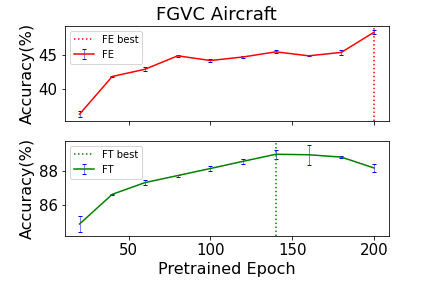}
    \includegraphics[width=0.32\linewidth]{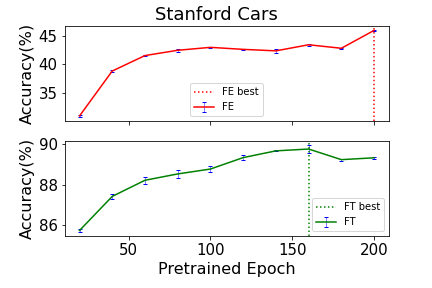}
    \includegraphics[width=0.32\linewidth]{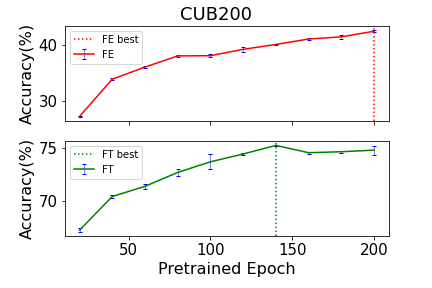}
    \includegraphics[width=0.32\linewidth]{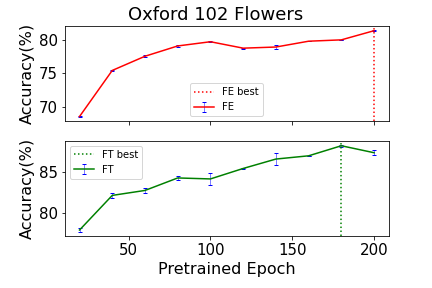}
    \includegraphics[width=0.32\linewidth]{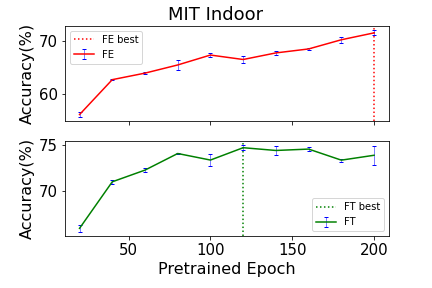}
    \includegraphics[width=0.32\linewidth]{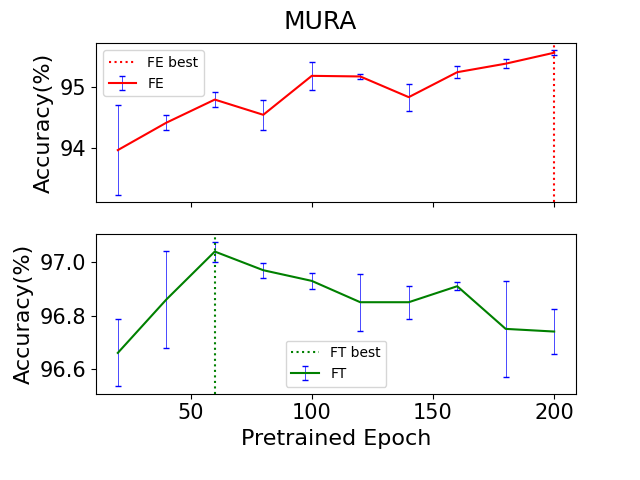}
    \caption{Transfer learning performance on selected  datasets with Moco-v2~\cite{chen2020improved}. Unlike supervised pre-training, the FE performance on self-supervised pre-training can be enhanced with more pre-training epochs.}
    \label{fig:ssl_exp}
\end{figure*}

Better performance in source tasks has long been believed to be more beneficial in target tasks. However, in this paper, we find that when using pre-trained models as feature extractors and retraining a new softmax classifier, the transferring performance does not agree with the source accuracy. There always exists the best epoch in the pre-training process. Intuitively, this is possibly brought by the distribution gap between the source and target data, 
forming a trade-off between source and target knowledge. If pre-training is less, no sufficient (general) visual knowledge can be obtained and the feature is suboptimal, but negative transfer happens the other way around. Based on this observation, we can operate a more sophisticated checkpoint selection process when we need a good feature extractor trained from source data\cite{lopes2017pre}.

Moreover, the common sense that better initialization should bring better training results is challenged given our observations. As can be seen from the difference in the evolution along the pre-training epochs between FE (view pre-trained model as a feature extractor) and FT (fine-tune the whole model), the FT performance still has a high correlation with the source performance, regardless of the U-property of FE performance. This means that a better feature extractor, which can be viewed as a better model initialization, does not definitely brings a better fine-tuning result. Further, in order to provide a more insightful explanation, we conduct a comparative experiment between the best FE model and the 
fully pre-trained one. Specifically, we delve into the spectral components of the feature before the classification layer and find that the components from top singular values contribute most to the FE, while the components with small singular values play a more critical role in the FT performance.
In previous research\cite{chen2019catastrophic}, spectral components corresponding to small singular values are criticized as hard to transfer or even untransferrable. Concretely, we reach consistent conclusions but take different operations. Unlike \cite{chen2019catastrophic}, we do not drop the residual component, but investigate its discriminativeness along with the main component. In this way, we empirically reveal the reason behind the paradox phenomenon that a better feature extractor fails to produce better fine-tuning results in the end. Consistent with an intuitive assumption that over-pre-training would undermine the performance of the pre-trained model as a feature extractor, we discover the main component of the target feature becomes impaired due to the model overfitting to source data with the pre-training epoch increasing. 

From a different perspective, we regard the main components as containing low-frequency knowledge of the feature, and the residual components as the carrier of high-frequency information. This makes sense since the residual components are generated from smaller 20$\%$ singular values, which are of high variation. In this way, our discoveries are also consistent with what has been well studied in the training mechanism of deep neural networks that the deep models learn low-frequency components before capturing high-frequency ones\cite{xu2019frequency,xu2019training}. 

However, there are still some phenomena beyond our explanation in Table~\ref{svd_cifar_mnist}. For example, since the performance of $\boldsymbol{F_m}$ decreases with more pre-training (from 88.24$\%$ to 58.74$\%$), what makes it grow much faster (40.34$\%$ vs. 11.02$\%$), though the accuracy is a little bit lower (99.08$\%$ vs. 99.30$\%$) when trained with target data? It is attractive to keep investigating the correspondence between different spectral components and different learning stages (e.g., early or late in pre-training, pre-training, or fine-tuning). We believe such research is beneficial for designing new regularizers for better transfer learning. Meanwhile, new assessment tools should be developed in the future since recent advanced methods cannot precisely select the best pre-training checkpoint during the same pre-training process.

Furthermore, how such mechanism work in self-supervised learning is also an interesting topic. We provide the results of self-supervised pre-training with MoCo-v2~\cite{chen2020improved} in Figure~\ref{fig:ssl_exp}. The results illustrate that, for FE, self-supervised pre-training does not obey the rules in the supervised case. We hypothesize that self-supervision, which is operated without explicit labels, alleviates the domain gap between source and target since it focuses more on learning an invariant mapping within the same training sample. Due to the page limit, we will investigate this difference between the two pre-training paradigms in future work.

\noindent
\textbf{Acknowledgements.} This research was supported by National Natural Science Foundation of China (NO.62106272), the Young Elite Scientists Sponsorship Program by CAST (2021QNRC001), in part by the Research Funds of Renmin University of China (NO. 21XNLG17)  and Public Computing Cloud, Renmin University of China.

{\small
\bibliographystyle{ieee_fullname}
\bibliography{inadequate}
}

\appendix
\section{Datasets}
\textbf{CIFAR10} \cite{krizhevsky2009learning} and \textbf{CIFAR100} \cite{krizhevsky2009learning} are two fundamental datasets in computer vision community. Both of them contain 50,000 training samples and 10,000 test samples, and all the samples are evenly distributed in each category. CIFAR10 consists of 10 common classes of objects. CIFAR100 includes 10 superclasses and each superclasses is made up of 10 fine-grained categories, and the size of each sample is 32 $\times$ 32.
\textbf{Food-101} \cite{bossard14} is a challenging food classification dataset, which consists of 101 categories. There are 250 clean test images for each class and 750 training images containing some noisy labels.
\textbf{FGVC Aircraft} \cite{maji2013fine} is a fine-grained dataset for aircraft classification. It contains 10,000 images of 100 categories of aircraft, and the training set is 2/3 of the whole dataset.
\textbf{Stanford Cars} \cite{stanford_cars} contains 196 classes of fine-grained cars, and there are 8,144 and 8,041 samples in the training set and test set, respectively.
\textbf{CUB-200-2011} \cite{WahCUB_200_2011} is a fine-grained bird classification dataset containing 200 species. There are 11,788 training samples and 5,894 test samples. Annotation of the bounding box, rough segmentation, and attributes are provided.
\textbf{Oxford 102 Flowers} \cite{nilsback2008automated} contains 200 common species of flowers in United Kingdom. Each of the categories has 40 up to 258 images. There are 2,040 training samples as well as 6,149 test samples.
\textbf{MIT Indoor 67} \cite{quattoni2009recognizing} contains 67 indoor scene categories with in total of 15,620 images, and 80$\%$ images are used for training.
\textbf{MURA}~\cite{rajpurkar2017mura} is a dataset of musculoskeletal radiographs, containing 40,561 X-ray images from 14,863 patient studies. The goal is to distinguish normal musculoskeletal examples from abnormal ones. We follow the common  setting to perform binary classification on each image. 

\section{Experimental setting of ResNet50}

\paragraph{Pre-training}
We borrow the official PyTorch implementation for ImageNet training using ResNet50. The total number of training epochs is set to 90. Stochastic gradient descent with a momentum of 0.9 is used to update the model parameters. The initial learning rate is 0.1 and is multiplied by 0.1 every 30 epochs. The weight decay is 1e-4. The pre-training performance is shown in Table~\ref{tab:pre-train}.

\begin{table}[h]
\centering
\caption{Pre-training performance of ResNet50 on ImageNet.}
\label{tab:pre-train}
\scalebox{0.75}{
\begin{tabular}{c|cccccccc}
\toprule
Epoch & 20 & 30 & 40 & 50 & 60 & 70 & 80 & 90 \\ 
\midrule
Acc$\%$ & 52.38 & 55.13 & 70.60 & 70.44 & 70.78 & 75.63 & 75.76 & 76.06\\
\bottomrule
\end{tabular}}
\end{table}
\vspace{-5mm}

\paragraph{Transfer learning}
In transfer learning, we use different training configurations to adapt to different datasets. For CIFAR10 and CIFAR100, in both FE and FT, the total training epoch is set as 150. The initial learning rate is 0.1 and is decayed by 10 times every 50 epochs. The optimizer is Adam\cite{kingma2014adam}. For the rest natural datasets, we run 6,000 iterations and 9,000 iterations for FE and FT, respectively; the learning rate is set to 0.1 for FT and 0.01 for FE. 

\section{Experimental setting of T2T-ViT\_t-14}

\paragraph{Pre-training}
We perform pre-training following the official codes of T2T-ViT~\cite{yuan2021tokens}. Specifically, we train T2T-ViT\_t-14 on ImageNet for 300 epochs. The final model achieves a Top-1 accuracy of 81.55\% on the ImageNet validation set. We choose checkpoints on epoch [20,40,60,80,100,120,150,200,250,300] for transfer learning experiments. 
\paragraph{Transfer learning}
For transfer learning, we perform the same data processing pipeline as used in ResNet50. For sufficient adaptation, the initial learning rate is set to 0.05 and decayed by a cosine annealing strategy, as suggested by the T2T-ViT paper. 

\section{Additional results}
Additional FE evaluation results on DTD~\cite{cimpoi2014describing} and Caltech256~\cite{griffin2007caltech} are shown in table~\ref{tab:add_FE}, which are also consistent with our claims. To further consolidate our conclusion, we also add the FE results of Swin-T in Table~\ref{tab:swin}. We train Swin-T for 300 epochs following the default settings in the official repository. The results of Swin-T clearly validate our conclusion that the best feature extractors are those inadequately pre-trained models.

\begin{table}[h]
\centering
\caption{FE evaluation of DTD and Caltech256.}
\label{tab:add_FE}
\scalebox{0.75}{
\begin{tabular}{c|cccccc}
\toprule
Pre-training Epoch &40 & 50 & 60 & 70 & 80 & 90 \\ 
\midrule
DTD ($\%$) & 69.36 & 69.15 & 69.36 & \textbf{70.43} & 68.88 & 69.84 \\
Caltech256 ($\%$) & 79.58 & 80.06 & 79.58 & 80.92 & \textbf{81.29} & 80.89 \\

\bottomrule
\end{tabular}}
\end{table}

\begin{table}[h]
\centering
\caption{Pre-training and FE performance on Swin-T.}
\label{tab:swin}
\scalebox{0.75}{
\begin{tabular}{c|ccccccc}
\toprule
Epoch & 240 & 250 & 260 & 270 & 280 & 290 & 300 \\ 
\midrule
pre-train  $\%$ & 79.89 & 80.20 & 80.51 & 80.88 & 81.00 & 81.10 & \textbf{81.21} \\ \hline
CIFAR100  $\%$ & 74.95 & \textbf{75.72} & 75.30 & 75.02 & 75.43 & 75.33 & 75.58 \\
DTD       $\%$ & \textbf{68.99} & 68.46 & 68.51 & 68.88 & 68.83 & 68.88 & 68.46 \\
Flower102 $\%$ & \textbf{86.39} & 86.24 & 85.95 & 85.93 & 85.75 & 86.37 & 85.61 \\
Aircraft  $\%$ & 44.47 & 45.47 & 43.97 & \textbf{45.92} & 44.69 & 45.53 & 43.94 \\
Indoor    $\%$ & 79.81 & 79.28 & 80.31 & \textbf{81.76} & 80.19 & 81.26 & 80.04  \\

\bottomrule
\end{tabular}}
\end{table}

\end{document}